\documentclass{article}

\usepackage[preprint]{neurips_data_2024}

\usepackage[utf8]{inputenc} % allow utf-8 input
\usepackage[T1]{fontenc}    % use 8-bit T1 fonts
\usepackage[table,xcdraw]{xcolor}
\definecolor{mydarkblue}{rgb}{0,0.08,0.65}
\usepackage[colorlinks=true,
    linkcolor=mydarkblue,
    citecolor=mydarkblue,
    filecolor=mydarkblue,
    urlcolor=mydarkblue]{hyperref}       % hyperlinks
\usepackage{url}            % simple URL typesetting
\usepackage{booktabs}       % professional-quality tables
\usepackage{amsfonts}       % blackboard math symbols
\usepackage{nicefrac}       % compact symbols for 1/2, etc.
\usepackage{microtype}      % microtypography
\usepackage{subcaption}
\usepackage{listings}
\usepackage{amsmath}
\usepackage{pifont}
\usepackage{graphicx}
\usepackage{amssymb}
\usepackage[title]{appendix}
\usepackage{cleveref}
\usepackage{soul}
\usepackage[shortlabels]{enumitem}
\usepackage{url}
\usepackage{listings}
\usepackage{color}
\usepackage{tikz}
\usepackage{balance}
\usepackage{multirow}
\usepackage{comment}
\usepackage{amsmath}
\usepackage{graphicx}
\usepackage{svg}
\usepackage{wrapfig,lipsum,booktabs}
\usepackage{titlesec}
\usepackage[frozencache,cachedir=.]{minted}
\usepackage[symbol]{footmisc}
\usepackage{tablefootnote}
\usepackage{dblfloatfix}

\newcommand{\code}{\texttt}

\definecolor{codegreen}{rgb}{0,0.6,0}
\definecolor{codegray}{rgb}{0.5,0.5,0.5}
\definecolor{codepurple}{rgb}{0.58,0,0.82}
\definecolor{backcolour}{rgb}{0.95,0.95,0.92}

\usepackage{subcaption}
\usepackage{graphicx}

\title{Zyda: A 1.3T Dataset for Open Language Modeling}

\author{%
  Yury Tokpanov$^*$ \\
  Zyphra \\
  \And
  Beren Millidge\thanks{ Core contributors.\\ Contact:\{yury, beren, paolo, jonathan, adam, james, quentin\}@zyphra.com}$^*$ \\
  Zyphra \\
  \And
  Paolo Glorioso \\
  Zyphra \\
  \And
  Jonathan Pilault \\
  Zyphra \\
  \AND
  Adam Ibrahim \\
  Zyphra \\
  \And
  James Whittington \\
  Zyphra \\
  \And
  Quentin Anthony \\
  Zyphra \\
}

\begin{document}

\maketitle

\begin{abstract}
%As large language models have been scaled up orders of magnitude over the last few years, their compute and data needs have grown enormously. State of the art language models trained today, even small ones, are typically trained for at least a trillion tokens and often significantly more. This rate of progress has outpaced the development of open-source freely accessible datasets for large-scale LLM pretraining. In this paper, we present Zyda (Zyphra Dataset), a permissively licensed dataset of 1.3 trillion tokens, created by combining all major existing well-respected open-source datasets into a high-quality dataset. We perform extensive and thorough filtering and deduplication (both intra- and inter-dataset) to ensure that the quality of the dataset is preserved and enhanced from its component datasets. We demonstrate that Zyda performs strongly against other open datasets such as Dolma and RefinedWeb, while significantly outperforming matched models of the Pythia suite and, due to our data processing pipeline, significantly outperforms all of its subsets alone.

The size of large language models (LLMs) has scaled dramatically in recent years and their computational and data requirements have surged correspondingly. State-of-the-art language models, even at relatively smaller sizes, typically require training on at least a trillion tokens. This rapid advancement has eclipsed the growth of open-source datasets available for large-scale LLM pretraining. In this paper, we introduce Zyda (Zyphra Dataset), a dataset under a permissive license comprising 1.3 trillion tokens, assembled by integrating several major respected open-source datasets into a single, high-quality corpus. We apply rigorous filtering and deduplication processes, both within and across datasets, to maintain and enhance the quality derived from the original datasets. Our evaluations show that Zyda not only competes favorably with other open datasets like Dolma, FineWeb, and RefinedWeb, but also substantially improves the performance of comparable models from the Pythia suite. Our rigorous data processing methods significantly enhance Zyda's effectiveness, outperforming even the best of its constituent datasets when used independently.

\end{abstract}

\section{Introduction}

Over the last five years, large language models (LLMs) have been undergoing an extremely rapid growth in scale, cost, and capabilities \citep{vaswani2017attention,radford2019language,brown2020language,team2023gemini,achiam2023gpt,sevilla2022compute}. 
%This has been driven by the discovery of underlying scaling laws which allow one to predict how a given amount of compute and data translate both to language modelling loss as well as, to some degree, downstream performance in other related tasks \citep{hestness2017deep,kaplan2020scaling,hoffmann2022training}. 
%The study of model behavior at various scale has allowed researchers to derive scaling laws to predict LLM loss and related task downstream performance based on the amount of compute and data used during pretraining \citep{hestness2017deep,kaplan2020scaling,hoffmann2022training}.
This development was fueled by the LLM scaling laws~\citep{hestness2017deep,kaplan2020scaling,hoffmann2022training} that established a relationship between the attainable loss and model size, dataset size and compute budget based on systematic experiments. Highlighting how performance increases with model size, these scaling laws provide guidance for how to optimally allocate resources for model size and dataset size given a fixed compute budget and provide concrete and fairly accurate predictions about the final loss and downstream capabilities of these models. 
% Such scaling laws posit that increased scale both in terms of parameter count as well as tokens trained upon uniformly tends to increase the performance of a model. 
% While initial scaling focused primarily on parameter counts \citep{kaplan2020scaling}, since the Chinchilla scaling laws \citep{hoffmann2022training} were released, focus has shifted significantly towards the scaling and improvement of datasets. 
Specifically, the ``Chinchilla scaling laws''~\citep{hoffmann2022training} suggest that equal scaling of parameters and data is required to train a ``compute-optimal'' model. However, as models become increasingly widely deployed, the majority of the total FLOPs are spent in inference, and not in pretraining. Therefore, the focus has been shifting towards ``inference-optimal'' models, which are much smaller and are trained on significantly more tokens than the Chinchilla scaling laws would recommend~\citep{touvron2023llama,jiang2023mistral}. 
These smaller models require significantly fewer forward-pass FLOPs and significantly less GPU VRAM for inference, which has caused them to become extremely important in the open-source LLM community, where the ability of a model to fit inside the VRAM of consumer GPUs is extremely important.

These trends have resulted in a significant increase in the ratio of total number of training tokens to model parameter count. State-of-the-art LLMs went from a 300B:175B ratio with GPT3 and a 540B:760B ratio with PaLM to a 12T:132B ratio with DBRX \citep{mosaic2024dbrx} and a 15T:8B ratio with Llama3. 
Research has also begun to focus strongly on dataset quality as a determinant of final model performance. 
While dataset quality has always been acknowledged as a factor, early models such as GPT3 \citep{brown2020language} were trained on largely unfiltered web-scraped data. 
Modern dataset processing techniques have become more advanced and rely on sophisticated pipelines of syntactic filtering, deduplication, and potentially semantic clustering and classification before a final dataset is produced. 
It has been consistently demonstrated that both filtering datasets \citep{rae2021scaling,elazar2023whats,raffel2020exploring} based on heuristic quality signals as well as aggressively deduplicating datasets \citep{lee2021deduplicating} offer significant improvements to the per-token average loss decrease during training and to model performance for a fixed compute budget. 
Beyond this, using more advanced methods such as utilizing semantic clustering to eliminate semantically duplicated data \citep{abbas2023semdedup,tirumala2024d4}, using trained classifiers to rank quality \citep{xie2023data,ilyas2022datamodels}, and discovering specific useful data subsets \citep{xie2024doremi} have been investigated to further improve model quality for a fixed data and compute budget. %As such, it appears that a significant determinant of the quality of a pretrained language model is the quality and quantity of the data upon which it was trained.

Open-source models have, to some extent, kept up with the frontier in performance during this period of unprecedented scaling. While these models \citep{touvron2023llama,team2024gemma,jiang2023mistral} are almost always 'open-weights', they usually give little to no information about the datasets they are trained on.
In part due to this, openly available datasets have tended to lag significantly behind the frontier in both scale and quality. 
%While there exist well-known open-source datasets in the 300B - 600B token range such as the Pile, SlimPajama, and RefinedWeb, no open, easily accessible, and permissively licensed teratoken scale datasets exist that can be used as an off-the-shelf dataset for teams training LLMs at this scale.
%, and manually collecting and processing a dataset for multi-trillion token training is highly nontrivial. 
If open models are to become competitive with the state of the art, they will need extremely large, general, and high-quality open datasets which can be utilized straightforwardly to train such models, allowing practitioners to focus on other aspects of performance such as model scaling and architecture.
Moreover, large high-quality datasets may encourage dataset standardization which can allow for fair and meaningful comparisons between different potential architectures and training methods. 

In this paper, we take a step towards this vision by releasing Zyda -- an open, permissively licensed dataset of 1.3 trillion tokens. Zyda was created by combining major permissively licensed open datasets which are recognized as high-quality within the community. Beyond simple collation, we performed extensive and thorough additional filtering and both intra- and inter-dataset deduplication on these datasets, in addition to any processing they originally underwent. Specifically, we design, test, and tune a novel filtering pipeline which is significantly more comprehensive than prior works. This pipeline utilizes a wide range of filters from a number of sources \citep{raffel2020exploring,slimpajama,kudugunta2024madlad,penedo2023refinedweb} and also introduces a number of novel ones, all of which were extensively manually tested and tuned to ensure efficacy. 

Building atop the deduplication pipeline from \citep{slimpajama}, we optimize it for highly parallel processing and carefully tune our deduplication hyperparameters with manual inspections to ensure a good false-positive rate. Inter-dataset deduplication is crucial, since simply combining existing datasets does not remove the duplicate data which is found in both datasets independently. We performed cross-dataset deduplication and found large numbers of cross-dataset duplicates, which is not surprising given that almost all existing open-source datasets were ultimately derived from Common Crawl\footnote{\url{https://commoncrawl.org/}} and filtered in similar ways. 

%We find that models trained on our dataset, Zyda, perform strongly on language modelling tasks -- significantly outperforming Dolma as well as equivalent Pythia models trained on the Pile. While, at the small scales we test, we find that the StarCoder dataset interferes with language modelling abilities, we also find that our dataset, Zyda, can achieve strong performance (see Fig. \ref{fig:dataset-comparison}, left panel) -- significantly outperforming RefinedWeb, its strongest subset, in equi-token comparisons while containing twice the token count thus enabling significantly better performance if the entire dataset is trained upon.
We find that models trained on Zyda perform strongly on language modelling tasks, significantly outperforming models trained upon Dolma as well as the Pile (see Fig.~\ref{fig:zyda-vs-external-datasets} left panel). Further, we find that StarCoder interferes with language modelling abilities, and when we remove the StarCoder subset, Zyda outperforms all its constituent datasets, including RefinedWeb which is known to be a particularly high quality dataset. While these comparisons are done at equi-token level, Zyda contains double the amount of tokens as RefinedWeb, thus enabling significantly better performance if the entire dataset is trained upon. We believe this improvement is based upon our postprocessing pipeline -- especially the heavy deduplication both within and across datasets. Zyda is openly available on Huggingface at \url{https://huggingface.co/datasets/Zyphra/Zyda} and all of our dataset processing code is open-source and is available at \url{https://github.com/Zyphra/Zyda_processing}.

\section{Dataset Composition and processing}

\subsection{Composition}
Our dataset comprises almost all currently accessible large-scale LLM pretraining datasets with permissive licenses. These include: The Pile \citep{gao2020pile}, SlimPajama \citep{slimpajama}, RefinedWeb \citep{penedo2023refinedweb}, C4 \citep{raffel2020exploring}, PeS2o \citep{peS2o}, \texttt{arxiv\_s2orc\_parsed} \citep{arxiv-s2orc-parsed}, and StarCoder \citep{li2023starcoder}. The Pile, SlimPajama, and RefinedWeb are general language modelling datasets which focus on text data and are primarily derived from Common Crawl and a few other sources. PeS2o is a dataset focused on scientific writing and is primarily comprised of arXiv and journal papers. StarCoder is a dataset focused on code and is comprised of code scraped and filtered from github.

\begin{figure}
    \centering
    \includegraphics[width=0.7\textwidth]{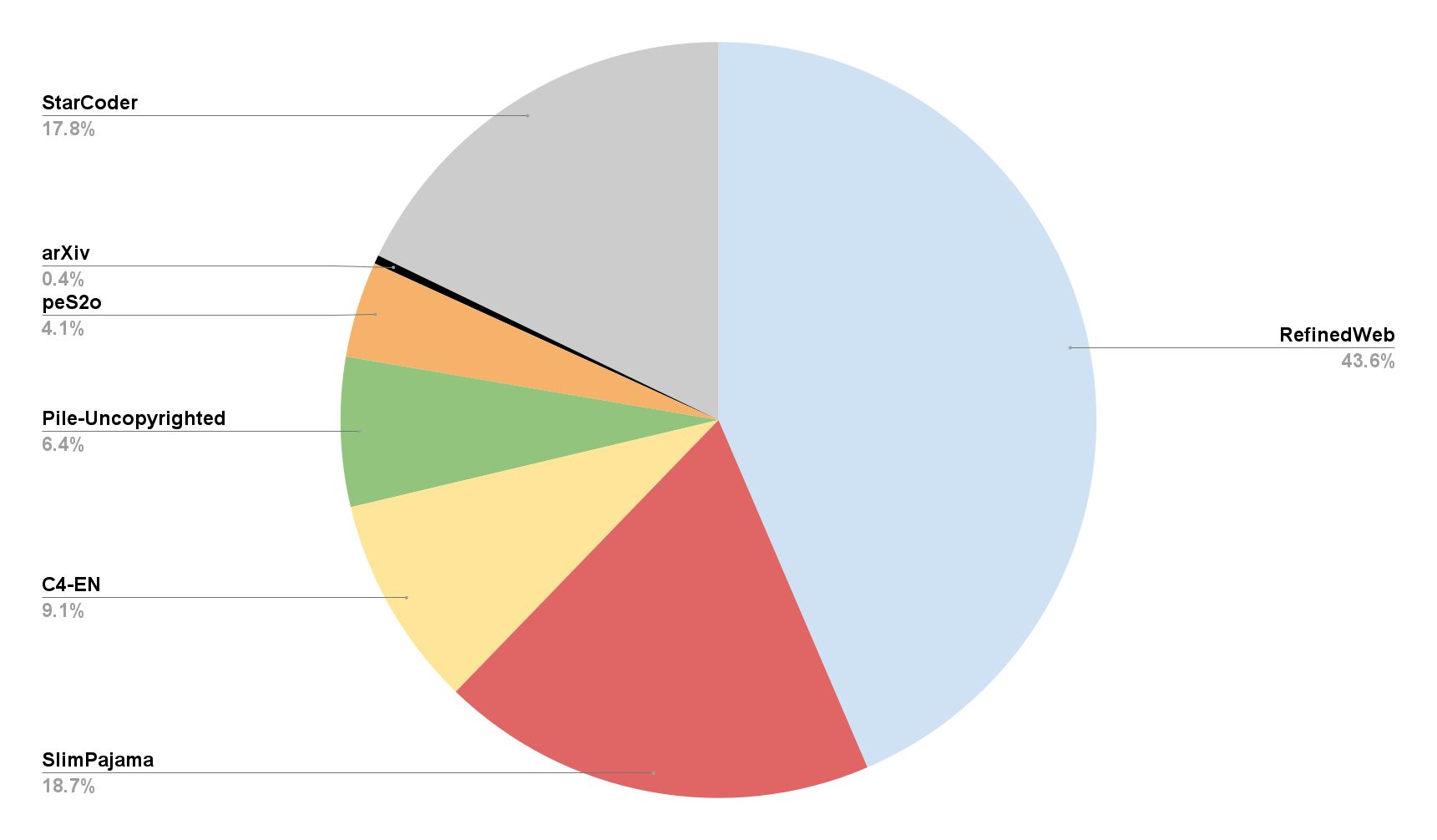}
    \caption{The proportion of different dataset subsets in Zyda. The primary proportion is RefinedWeb, followed by SlimPajama and StarCoder.}
    \vspace{-2ex}
    \label{fig:zyda-pie}
\end{figure}

\subsection{Filtering}

\begin{table*}[b]
\centering
\begin{tabular}{|>{\columncolor[HTML]{EFEFEF}}c|c|c|c|}
\hline
\cellcolor[HTML]{C0C0C0}\textbf{Dataset} & \multicolumn{1}{c|}{\cellcolor[HTML]{C0C0C0}\textbf{Rows Initial (Millions)}} & \multicolumn{1}{c|}{\cellcolor[HTML]{C0C0C0}\textbf{Rows Removed (Millions)}} & \multicolumn{1}{c|}{\cellcolor[HTML]{C0C0C0}\textbf{Percent filtered}} \\ \hline
RefinedWeb & 968.000 & 21.350 & 2.21\% \\ \hline
SlimPajama & 590.395 & 21.045 & 3.56\% \\ \hline
C4-EN & 364.869 & 4.234 & 1.16\% \\ \hline 
Pile-Uncopyrighted & 177.010 & 21.536 & 12.17\% \\ \hline 
peS2o & 38.811 & 0.044 & 0.11\% \\ \hline
arXiv & 1.672 & 0.189 & 11.33\% \\ \hline
\end{tabular}
%\vspace{0.2cm}
\caption{\small Number of document and percentage of all documents removed from each subcomponent dataset by our filtering process.}
\label{tab:filtering_removed}
\end{table*}

Prior to deduplication, following common practice \citep{gao2020pile,rae2021scaling}, we performed heuristic syntactic filtering for quality and to remove low-quality data, such as meaningless strings, large quantity of random numbers, as well as pornographic or otherwise objectionable content. Our filtering pipeline consisted of two stages: (1) substring replacement and (2) document-level filtering and removal.

For the first stage, we deployed regexes that replace certain substrings with more sanitized versions. This was primarily to fix common formatting issues that we noticed in our original analyses of the datasets. Examples included excessively long sequences of dashes, full-stops, `\textbackslash r' characters, and other punctuation characters which presumably arose from quirks of formatting or the processing pipelines that lead to these documents. Such strings appeared with reasonable frequency and often were found amid otherwise unobjectionable documents, so we did not wish to simply remove documents in which they appeared. Typically, we replaced large numbers of repeated characters with a single or just a few characters. 
For instance, we replaced large numbers of linebreaks `\textbackslash n' and carriage returns `\textbackslash r' with just a single linebreak or carriage return. Similarly, for long sequences of dashes, we replaced them with a a single dash.
For the second stage, we performed document-level filtering based on a set of syntactic heuristics which can be cheaply computed from the raw text of a document.
If the threshold of the filter was exceeded, the whole document was removed from the dataset. These filters broadly fell into three categories: (1) removing syntactically broken or otherwise gibberish documents, (2) removing semantically meaningless documents, and (3) removing meaningful but objectionable content. Examples of the first kind of filter include methods like filtering based on the proportion of alphanumeric characters in a document, which at high proportions uniformly corresponds to documents comprised of entirely gibberish strings generated either through broken preprocessing or through unknown processes on the internet. Semantically meaningless documents include documents full of seemingly random numbers, cryptographic strings, and lists of seemingly unrelated URLs. Objectionable content included primarily pornographic and offensive content, which we removed with specialized word lists.

Our general philosophy in filtering was to not filter excessively and keep false positives (i.e. good documents removed) relatively low. We manually tested and tuned each of the filters presented here on the Pile dataset. We tuned the filter thresholds so that we obtained a false-positive threshold of about 20\% – that is, 20\% of the filtered documents were seemingly unobjectionable, while 80\% were obviously harmful and it was correct to remove them. This provides a reasonable trade-off between excessive filtering while still removing the majority of content the filter was aimed at reducing. For every filter, we spent significant time manually tuning the threshold and other parameters, then looking at the filtered outputs and attempting to identify true vs false positives. We primarily performed this tuning on the Pile dataset and performed only sanity checking of the filter outputs on the other datasets. However, we believe that many of the categories of low-quality text likely have similar distributions between datasets, since so many of them are derived from the same source. A full list and description of all filters can be found in Appendix \ref{Filtering_details}. 
All datasets were filtered using the same rules except for StarCoder, which we exempted since it consists entirely of code which has a significantly different distribution than the primarily text data in the other datasets. Additionally, the StarCoder authors performed a thorough code-specific filtering of the dataset before release. Table \ref{tab:filtering_removed} shows the proportion and number of rows removed from each dataset by our filtering process. We observe that the primary source of data removed by our filters is the Pile and arXiv.

\subsection{Deduplication}

\begin{figure}[htbp]
    \centering
    \subcaptionbox{Distribution of edit similarity distance of selected duplicate documents, red dash line marks 40\% threshold.\label{fig:distr_es}}
        {\includegraphics[width=.50\textwidth]{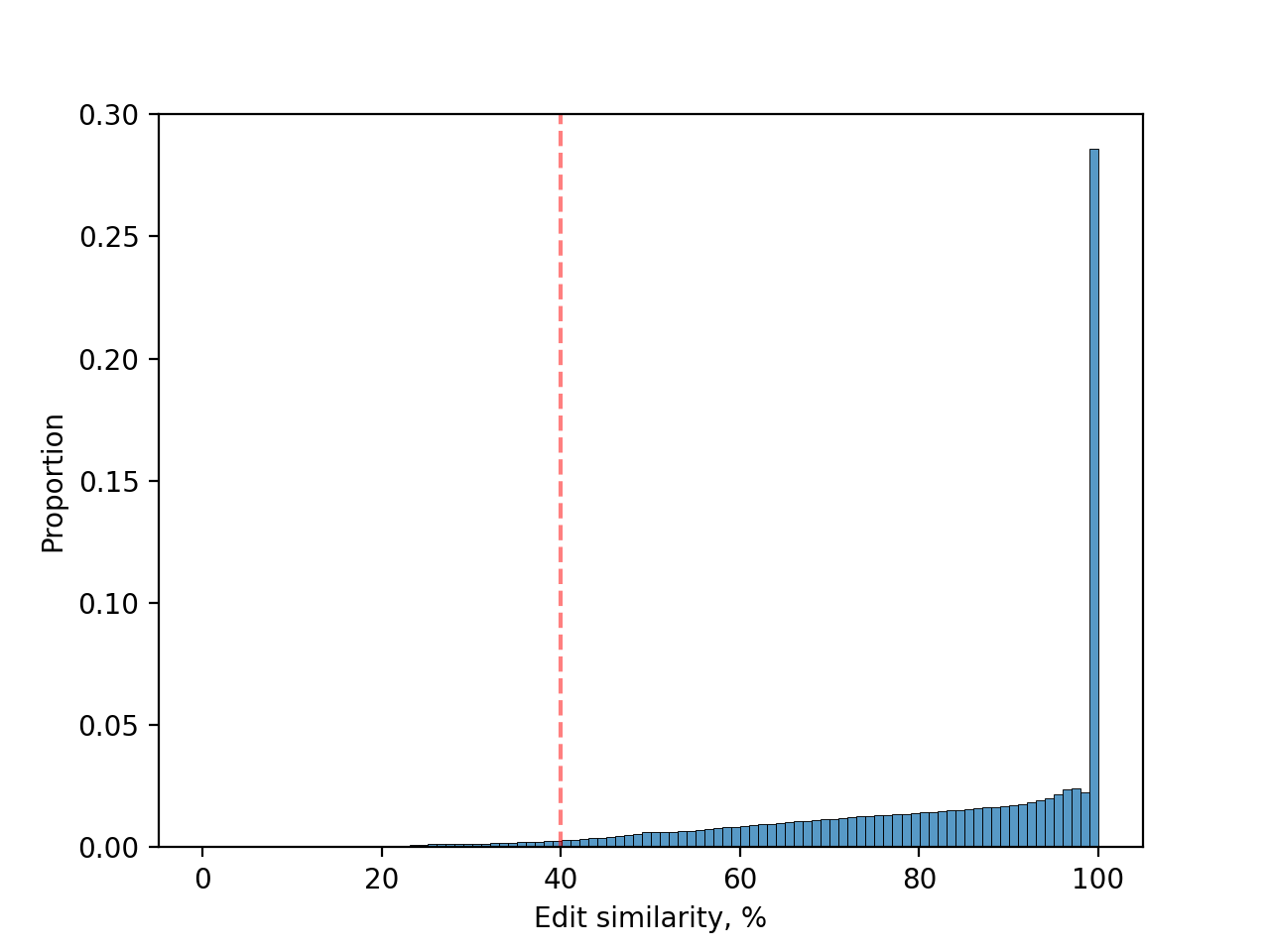}}
    \hfill
    \subcaptionbox{Distribution of Jaccard similarity distance of selected duplicate documents, red dash line marks 40\% threshold.\label{fig:distr_js}}
        {\includegraphics[width=.45\textwidth]{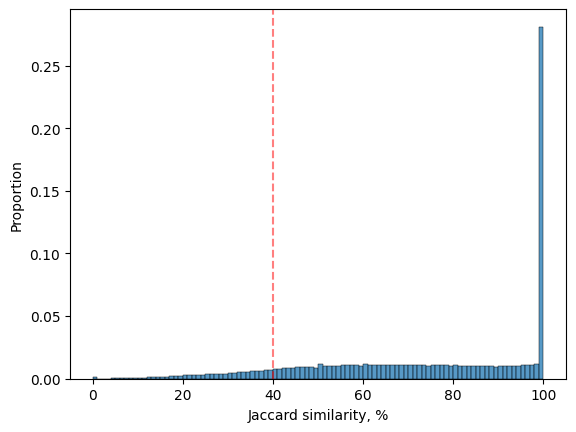}}
    \caption{Document similarity distances for Zyda dataset}
    \vspace{-2ex}
\end{figure}

\begin{table*}[b]
\vspace{-4ex}
\centering
\begin{tabular}{|>{\columncolor[HTML]{EFEFEF}}c|c|c|c|}
\hline
\cellcolor[HTML]{C0C0C0}\textbf{Dataset} & \cellcolor[HTML]{C0C0C0}\textbf{Raw Tokens} & \cellcolor[HTML]{C0C0C0}\textbf{LSH 80\% tokens} & \cellcolor[HTML]{C0C0C0}\textbf{LSH 40\% tokens} \\ \hline
RefinedWeb & 579,958,873,663 & 569,412,954,030 & 564,793,074,285 \\ \hline
SlimPajama & 605,453,814,327 & 355,300,814,841 & 242,280,662,390 \\ \hline
C4-en & 174,402,215,716 & 156,848,709,800 & 117,511,570,406 \\ \hline
Pile-Uncopyrighted & 253,107,739,551 & 115,280,920,159 & 82,918,673,353 \\ \hline
peS2o & 57,171,106,720 & 56,264,262,881 & 53,376,291,935 \\ \hline
arXiv & 24,261,321,215 & 19,523,122,996 & 4,703,637,744 \\ \hline
StarCoder & 260,865,965,654 & 257,327,664,673 & 231,285,458,516 \\ \hline
\textbf{Total:} & \textbf{1,955,221,036,846} & \textbf{1,529,958,449,380} & \textbf{1,296,869,368,629} \\ \hline
\end{tabular}
\caption{\small Initial and final number of tokens before and after filtering and deduplication with two thresholds (80\% and 40\%)}
\vspace{0.2cm}
\label{tab:deduplication_token_counts}
\end{table*}
To identify duplicates we used Locality Sensitive Hashing (LSH) based on MinHash signatures~\citep{broder1997minhash}. 
This technique allows fast approximate identification of duplicate candidates based on Jaccard similarity of sets of $n$-grams $S_n(\cdot)$ in documents. For example, using $\text{MinHash}(A) = \{ \min h_1(S_n(A)), \ldots, \min h_k(S_n(A)) \}$ and $\text{MinHash}(B) = \{ \min h_1(S_n(B)), \ldots, \min h_k(S_B) \}$ for $k$ hash functions, we can measure the resemblance of documents $A$ and $B$ using:

\begin{equation}
\label{eq:resemblance}
\frac{1}{k} \sum_{i=1}^k \mathbf{1}[\min h_i(S_n(A)) = \min h_i(S_n(B))].
\end{equation}

In the rest of the paper, we refer to LSH-$x$\% as the LSH with at least $x$\% resemblance, as expressed by Equation \ref{eq:resemblance}. We deduplicated each dataset both against itself and against the other datasets in our full dataset. We built the LSH index by inserting the components in the following order: first Pile-uncopyrighted, then C4-en, peS2o, \texttt{arxiv\_s2orc\_parsed}, RefinedWeb, SlimPajama, and finally StarCoder.

For our deduplication pipeline, we used 13-grams\footnote{We chose 13-gram based on what \citep{gao2020pile, soldaini2024dolma} use, which is a common choice of $n$-gram. Other choices of $n$ can be successfully applied to deduplicate text. For example, \citep{penedo2023refinedweb, penedo2024fineweb} successfully use 5-grams for deduplication} based on words to form our n-gram subsets and a minhash signature size of 128. Before generating 13-grams, we performed NFC normalization, conversion to lower case, and removal of punctuation and consecutive spaces, newlines, tabs in the middle and in the beginning and end of the strings. We performed deduplication at two Jaccard similarity thresholds: 40\% and 80\% (documents are considered duplicates if their similarity measure is equal or greater than the threshold). The parameters of LSH index were optimized to minimize the rates of false positives (FP) and false negatives (FN): for the 40\% threshold, minhash indices were split into 32 bands each with a range of 4, while for the 80\% threshold, we used 9 bands with a range of 13. From these parameters we can derive the following false-positive and false-negative rates: for 40\% threshold, the false-negative rate is $3.4\%$ and false-positive rate is $5.4\%$, for 80\% threshold, the false-negative rate is $3.3\%$ and false-positive rate is $2.5\%$. Table \ref{tab:deduplication_token_counts} summarizes how many tokens (using gpt-neox tokenizer) were removed at different thresholds. Our raw dataset consists of 2T tokens, and we end up with 1.5T for 80\% threshold and 1.3T for 40\% threshold.

After identifying duplicate pairs using the LSH minhash technique, we clustered documents into a graph of connected components with the nodes being documents and the edges connecting duplicate pairs. We then kept only one document from the cluster while removing the rest. To determine which document to keep, we sorted the documents in the clusters by their dataset of origin and kept the highest-ranking one according to the following order: 1) StarCoder; 2) RefinedWeb; 3) peS2o; 4) arXiv; 5) C4; 6) Pile-uncopyrighted; 7) SlimPajama. We chose StarCoder as the highest-ranking dataset because it was specifically designed for code, so we hypothesized that any duplicate code snippets are likely to be of the highest quality from this source. We chose the rest of the ranking based on heuristic assessments of quality.

We then performed random sampling of duplicates in the clusters to manually explore examples. The largest clusters usually contained either short low-quality documents or documents with widely distributed texts, such as license agreements, advertisements, etc. We did notice that at 40\% threshold LSH minhash was performing a qualitatively different kind of deduplication than at 80\%: while at 80\% most duplicates looked very similar, at 40\% we started seeing duplicates across formats, especially between peS2o and arXiv components of our dataset, where we observed, for instance, the same paper but formatted in two different ways.

Since the LSH minhash algorithm only performs approximate deduplication, we also sampled 4.8 million duplicate pairs to estimate the actual false-positive rate based on Jaccard and edit similarities. We define the {\it edit similarity} between two documents as their edit distance divided by maximum length of two documents. We found good agreements between theoretical and estimated false-positive rate based on Jaccard similarities, while the estimated false-positive rate based on edit similarity for the 40\% threshold was even lower than the theoretical estimate at $3.1\%$. Figures \ref{fig:distr_es} and \ref{fig:distr_js} show the distribution of edit and Jaccard similarities for the 40\% threshold version of our dataset: As expected, the vast majority of identified duplicates are above the threshold (marked as red dash line), and for the edit similarity metric, the distributions is skewed toward higher values (which is expected, since it corresponds to a lower false-positive rate). To compare the 40\% and the 80\% versions of Zyda, we trained 1.4B transformers on 50B tokens sampled from each version of Zyda. We found that 40\% Zyda performed slightly better (Fig. \ref{fig:dedup-ablations}). Due to this, we chose to release the 40\% version as our primary Zyda dataset.

\begin{comment}
\section{Dataset Analysis}
\begin{itemize}
    \item Histogram of document lengths
    \item Table of where deduplicated from which documents
    \item Document lengths removed by filtering
    \item Document lengths removed by dedupe
    \item Comparison of 80\% and 40\% threshold dedupe
    \item Comparison of edit distance vs jaccard similarity and false positive rates as a function of dataset length
    \item distribution of languages in the dataset (?)
    \item Examples of things removed by filters and dedupe
    \item Histogram of distributions of edit similarities of randomly selected dupes
    \item Anything else?
\end{itemize}
\end{comment}

\section{Performance}

\begin{figure}
    \centering
    \includegraphics[width=0.7\linewidth]{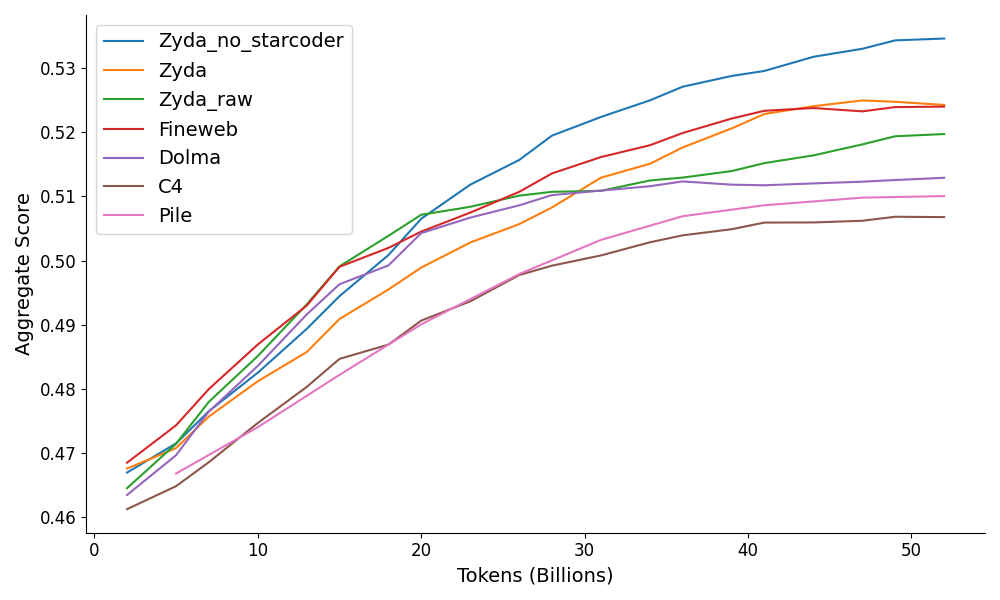}
    \caption{Aggregate evaluation scores across training steps for a 1.4B model trained on 50B tokens of Zyda and comparable datasets. Zyda and especially Zyda without starcoder outperform strong open baselines such as FineWeb, and Dolma. Aggregate scores is the mean of arc-challenge, arc-easy, boolq, openbookqa, piqa, sciq, and winogrande. Scores smoothed using a window size of 5.}
    \label{fig:scores-across-time}
    \vspace{-3ex}
\end{figure}

We perform a series of training runs of small transformer models to assess the performance of our dataset vs alternatives such as Pile, and RefinedWeb, and Dolma. We first compare the performance of models trained on our dataset with the well-known Pythia suite of models \citep{biderman2023pythia}. We match the architecture, hyperparameters, and training process of the Pythia suite. We compare against the 410M Pythia model size. Like Pythia, we train for 300B tokens at this model scale. This provides a direct apples-to-apples comparison of Zyda against the Pile. We demonstrate that across the range of evals reported in the Pythia paper, that Zyda-trained models significantly outperforms the Pile. We especially observe gains in reasoning evaluations such as ARC and PIQA.

For a fair comparison, we ensure that these comparisons are equi-token. We note that Zyda contains 1.3T tokens, while the Pile contains only 300B, and thus we expect that models trained on the full Zyda dataset would substantially outperform models trained on the full Pile, such as Pythia, even more than we do so here.

\begin{figure*}[htbp]
    \centering
    \subcaptionbox{Aggregate evaluation performance of models trained on Zyda versus Pile. Aggregate score is the mean of the evaluation scores for ARC-c, ARC-e, LogiQA, PIQA, SciQ, and WinoGrande
    \label{fig:zyda-vs-pile-across-scale}}
        {\includegraphics[width=.49\textwidth]{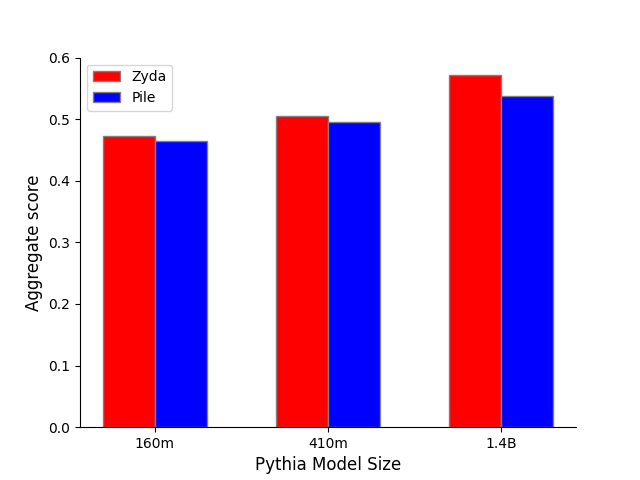}}
    \hfill
    \subcaptionbox{Evaluation scores for 1.4B Pythia trained on Zyda and Pile.
    \label{fig:zyda-vs-pile-across-evals}}
        {\includegraphics[width=.49\textwidth]{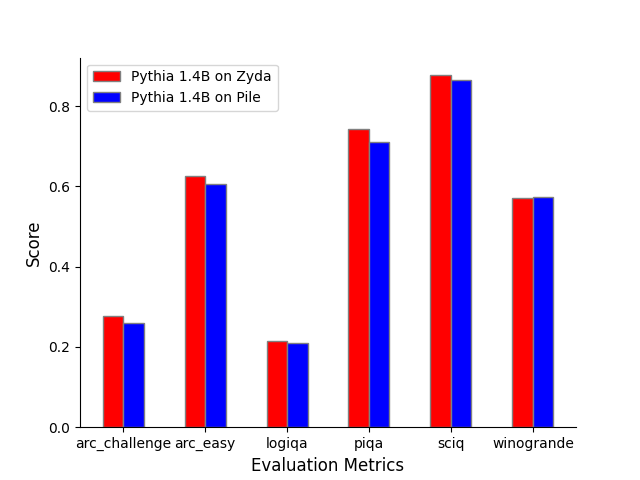}}
    \caption{We match the Pythia suite in architecture and training hyperparameters. We observe that Zyda outperforms Pile on evaluations and that this advantage increases with scale, which we believe to be due to reduced noise on standard evals as model performance improves. All models were trained for 300B tokens on either Pile or Zyda.}
    \vspace{-2ex}
\end{figure*}

\begin{figure*}[htbp]
    \centering
    \subcaptionbox{Comparison of Zyda performance vs other datasets such as Dolma, RefinedWeb, subsets of Zyda, and Zyda without any preprocessing (Zyda Raw). All comparisons done for 1.4B Pythia on 50B tokens.
    %, which is surprising since these are its subsets. 
    %We attribute the improved performance to our stringent filtering and deduplication pipeline. We find that Zyda significantly outperforms comparable open datasets such as Dolma.
    \label{fig:zyda-vs-external-datasets}}
        {\includegraphics[width=.45\textwidth]{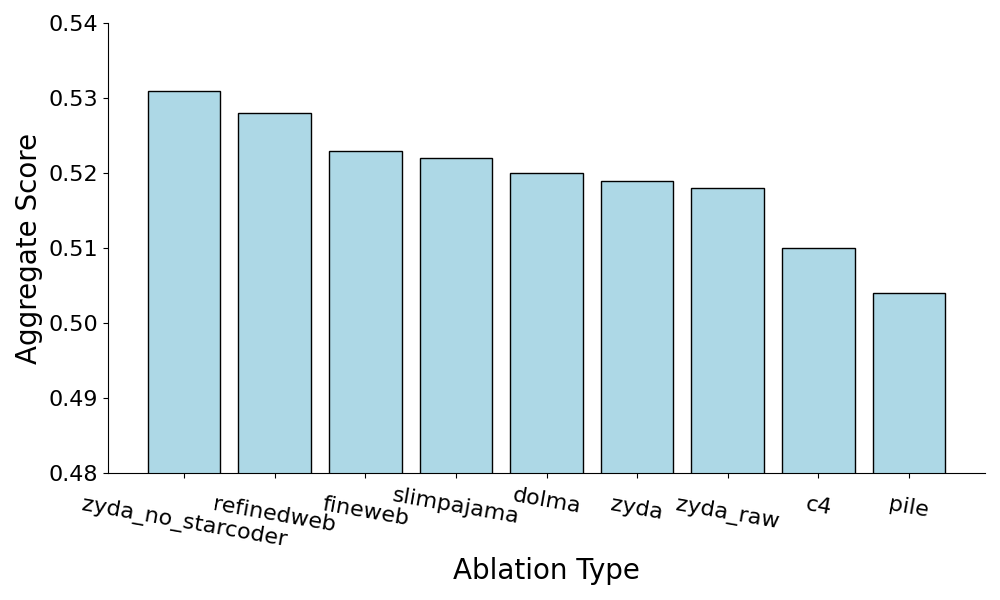}}
    \hfill
    \subcaptionbox{The effect of deduplication threshold at either 40\% or 80\% similarity. We observe a slight advantage for 40\% although this is not consistent across evaluation metrics.
    \label{fig:dedup-ablations}}
        {\includegraphics[width=.45\textwidth]{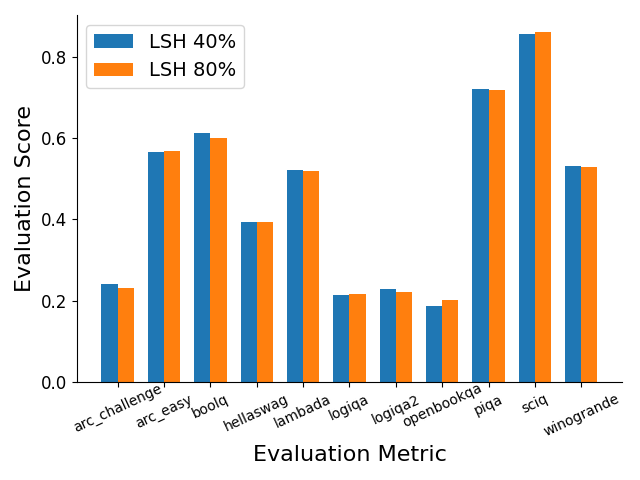}}
    \caption{Comparison of Zyda with alternative datasets, and across deduplication LSH}
    \vspace{-2ex}
\end{figure*}

Next we perform a preliminary scaling analysis of our dataset against the Pile and Pythia on a range of scales from 160M to 1.4B, all trained on the full Pile and 300B tokens of Zyda. We show that the advantage of Zyda appears to increase with scale. We suggest that this is because the increased dataset quality is not `visible' to small models as they are capacity limited to only extract the larger modes of variance in the data, whereas larger models can additionally `see' finer modes of variance where our increased dataset quality provides an advantage. A small model may simply be unable to fit important aspects of language regardless of the dataset used to train it.
%This is perhaps because larger models are able to perform significantly above chance at more evals, enabling a better signal of dataset quality to be derived.
%perhaps due to the larger scale enabling a less noisy signal of the higher quality of Zyda to become more obvious.
\begin{comment}
\begin{figure}
    \centering
    \includegraphics[width=0.9\linewidth]{figures/zyda_pythia_scaling_plot.png}
    \caption{Performance of Zyda against Pythia across scales at 300B tokens. We see the advantage for Zyda appears to grow with scale.}
    \label{fig:enter-label}
\end{figure}
\end{comment}
Finally, we performed a number of ablations of our dataset and comparisons to other datasets at the 50B tokens, 1.4B model scale. We plot the aggregate performance on a range of evaluations of final checkpoints on Figure \ref{fig:zyda-vs-external-datasets}, and performance as a function training steps on Figure \ref{fig:scores-across-time}.

We first observe that Zyda at the 50B token stage performs slightly worse than RefinedWeb but still better than SlimPajama C4, and Dolma. \emph{A priori}, this is not surprising since Zyda contains both of these as subsets. The performance decrease of Zyda (compared to RefinedWed) is attributed to the presence of StarCoder. With StarCoder removed, Zyda outperforms all other datasets and indeed performs better than any of its subsets -- a surprising result which speaks to the importance of our filtering and deduplication pipeline. It appears that, at least at this scale, the competition between language and code introduced by having a significant fraction of code inhibits performance on standard language modelling evaluations. We find that a comparable open dataset -- Dolma \cite{soldaini2024dolma} -- performs worse than both Zyda and most individual subsets of Zyda. It is also important to note that RefinedWeb only contains 600B tokens to Zyda's 1.3T and hence we anticipate Zyda would greatly outperform RefinedWeb if the model was trained to the end on each dataset. RefinedWeb is a strong and already well-filtered dataset which also forms a significant proportion of Zyda, and is thus a strong baseline. Zyda also contains a number of less performant datasets such as the Pile and C4. Despite this, Zyda's additional filtering and deduplication steps boost Zyda's performance such that the full dataset performs better than the raw version of its strongest subset. We can also detect this by noting that the raw Zyda dataset performs averagely at about the level of SlimPajama. Ultimately, Zyda appears to be a strong pretraining dataset at the 1T token scale containing only pretraining web data. Since it appears to decrease performance, we recommend removing StarCoder from Zyda for small models which need to be language focused instead of code focused since, at this scale, code appears to cause interference with the model's language abilities. The performance of Zyda has also been empirically validated through our language modelling effort Zamba \citep{glorioso2024zamba}, a 7B hybrid SSM-transformer which performs very strongly per pretraining token, and was trained on an early version of Zyda.

\section{Related Work}

Concurrently with the rise of open-source large language models, there has been an increase in the number of open-source and widely available datasets for large-scale language model pretraining. Early works in this vein include C4 \citep{raffel2020exploring}, derived directly from Common Crawl, and The Pile \citep{gao2020pile} which was manually constructed from a number of disparate sources. Other, more recently released large datasets such as SlimPajama \citep{slimpajama}, RefinedWeb \citep{penedo2023refinedweb} and Dolma \citep{soldaini2024dolma} have undergone more thorough filtering and deduplication -- a process which we build upon in this work. We have also built upon works that described their filtering pipelines without releasing the dataset. Of especial use was Gopher \citep{rae2021scaling} and RedPajama \citep{redpajama}, which inspired a number of our own filters. 

Most similar to our approach are RefinedWeb and Dolma. The RefinedWeb authors, as part of the Falcon LLM team, created a dataset of several trillion tokens using a filtering and deduplication pipeline similar to ours, but they only released publicly approximately 600B tokens of their full dataset. The Dolma dataset is also similar to our approach and releases several trillion tokens, derived from a mixture of Common Crawl and some existing datasets such as C4 and PeS2o, which are also included in our dataset. Hovewer, Dolma does not cross-deduplicate data between its component datasets -- a process we found likely to be important since we identified many duplicate documents across datasets. We additionally find that Zyda outperforms Dolma at equi-token language modelling evaluations by a significant margin. Concurrently, FineWeb~\citep{penedo2024fineweb} was also released, which offers 15T tokens directly from Common Crawl under an open license. Since their Common Crawl datasets appear to come from a different source, it seems likely that Zyda and FineWeb can be productively merged. Zyda consolidates all the primary prior open-datasets into one place and subjects them to a uniform quality filtering and deduplication which can then be augmented with FineWeb's additional Common Crawl tokens.

Additionally, unlike these works which directly filter from Common Crawl, we aim to instead collect and collate existing open-source datasets under one banner, and to uniformly filter and deduplicate them against one another so as to create a dataset of known high-quality text which can be used to perform training at trillion-token scales. We hope that this seed dataset can then be expanded as additional datasets are released by the open-source community. 

We built our deduplication pipeline upon SlimPajama's existing open-source libraries, although we ended up significantly modifying them and performed extensive manual tuning and optimization. Our initial filters were inspired by the filter list accompanying the RedPajama dataset. However, after preliminary testing, we removed some of their filters which we judged were ineffective and integrated new ones, either from other papers such as C4, Gopher, and MADLAD-400 \citep{kudugunta2024madlad}, or of our own invention. A full list and description of the filters we used, as well as qualitative impressions gained during testing them, can be found in Appendix~\ref{Filtering_details}.

%– We are unique in that we provide a unified dataset with additional cleaning and aggressive deduplication of all datasets against all other datasets resulting in a high quality multi-trillion token open dataset
%– This means that practitioners wanting to pretrain at the trillion scale don’t necessarily have to roll their own datasets anymore
%– The dataset idea can be extended and new datasets can be added as they are released creating a permanently growing open-source high quality dataset available for LLM pretrainers
%– In future work we could improve filtering, deduplication etc, integrate sources of synthetic data, improve semantic filtering etc using new techniques

\section{Discussion}

In this paper, we have presented Zyda, a unified dataset released under a permissive license, comprising most of the largest and highest quality existing open-source datasets available. Upon these, we have performed extensive additional filtering, beyond what was originally applied, in addition to thorough intra- and inter-dataset deduplication. Our aim with this work is to create a growing and extendable dataset which can be easily used by practitioners to train trillion-token scale language models while also consolidating and unifying the efforts made by disparate open-source groups who have released their datasets. Ultimately, we hope that our work can provide an ``off-the-shelf'' accessible trillion-scale high-quality pretraining dataset which can be used by groups aiming to pretrain their own LLMs.

While not performing new collection from Common Crawl, we believe that our work is an important and valuable step towards the creation of large scale high quality open datasets given that there exist a number of high quality existing datasets but few to none of them individually reach the scale necessary for training state-of-the-art models. Collating, filtering, and deduplicating the existing datasets needed to create a trillion token dataset is nontrivial work and extremely important to raise the quality of the dataset and prevent significant amounts of inter-dataset duplicates. This latter operation of deduplication between datasets is extremely important given the degree of duplicated documents we discovered in common open-source datasets.

Ultimately, we believe that this dataset is only the first step towards building an open dataset competitive with state-of-the-art models at the largest scales. While our dataset combines the primary filtered open-source datasets available, the datasets used to train the largest models today are already an order of magnitude larger. The recent release of FineWeb \citep{penedo2024fineweb} which contains 15T Common Crawl tokens is exciting and effectively removes the first bottleneck -- sourcing the tokens -- to the creation of large effective open datasets to LLM pretraining, yet likely the best proprietary datasets still retain significant quality advantages. While many of the secrets to high quality filtering remain hidden behind the veil of the largest labs, there is now a growing amount of academic work showing methods for how dataset quality can be significantly improved beyond the simple filtering and deduplication pipeline we have implemented here. We believe that a strong pretraining mixture can likely be created by merging Zyda and FineWeb together which can be used to approach the training token count of the largest closed datasets.

While we have performed thorough filtering and deduplication to remove low quality documents and many duplicates, we have only yet taken the first step in improving dataset quality. A large number of methods can still be employed to further improve its quality, at the cost of greater compute requirements. Such methods include training semantic classifiers to detect high or low quality data \citep{xie2023data,ilyas2022datamodels}, performing passes over the dataset with language models to filter based on the perplexity of the data \citep{marion2023less}, performing clustering on the dataset to remove outlier or repetitive data \citep{tirumala2024d4}, augmenting the original data with synthetic or rephrased data \citep{maini2024rephrasing}, and many other approaches. We hope that future work explores these and other avenues and continues to extend and improve the underlying open datasets available for pretraining so as to push forward the frontier of performance accessible with open datasets.

\clearpage

\bibliographystyle{apalike}
\bibliography{main}

\appendix
%\begin{appendices}

%\begin{comment}
\section*{Author Contributions}
\textbf{Yury} — Led dataset processing and deduplication pipeline. Led ablation and performance experiments. Contributed to paper writing, figures, and edits.

\textbf{Beren} — Led dataset filtering pipeline. Contributed to project conceptualization. Wrote primary draft of paper and contributed to figures and edits.

\textbf{Paolo} — Contributed to paper writing, figures, and edits.

\textbf{Jonathan} — Contributed to paper writing and edited the paper draft.

\textbf{Adam} — Contributed to the paper draft.

\textbf{James} — Contributed comments to the paper draft.

\textbf{Quentin} — Contributed to paper writing, creating figures, and edits. Contributed to project conceptualization.
%\end{comment}

\section{Limitations}

While we believe Zyda is a highly effective dataset for language modelling, and this is supported by our ablation studies, and our empirical successes with the Zamba model \citep{glorioso2024zamba}. However, except our ablations comparisons against the Pythia suite, all our other ablations were only trained on 50 billion tokens which may not be enough to predict dataset quality over the full course of training over one or more epochs on the dataset. We limited our ablations to 50B tokens on a 1.4B model primarily out of compute and cost considerations. Additionally, while we performed a small scaling suite for the Pythia models, we did not perform this for other ablations, focusing on the 1.4B scale. It is possible that the scaling properties of different datasets differ so tht some datasets may perform better with larger or smaller models, and we cannot discount that this could change the rank ordering of datasets by performance.

\section{Ablation experimental details}

For our 50B ablations, we used a 1.4B standard transformer model. The architecture and trainng hyperparameters are presented in Table \ref{ablations_hyperparam_table}. We used the Swiglu activation function on the MLP layers and trained with RoPE position embeddings. 

\begin{table}[H]
\begin{tabular}{
>{\columncolor[HTML]{EFEFEF}}c c}
\cellcolor[HTML]{C0C0C0}\textbf{Hyperparameter} & \cellcolor[HTML]{C0C0C0}\textbf{Value} \\
Number of Layers                                & 24                                     \\
Hidden Dimension                                & 248                                   \\

Number of Attention Heads                       & 16                                     \\
Batch Size                                      & 512                                    \\
Max Learning Rage                               & 2.0e-4                                 \\
LR Decay Schedule                               & Cosine                                 \\
Minimum LR                                      & 2.0e-5                                 \\
Weight Decay                                    & 0.1                                    \\
Adam Beta2                                      & 0.95                                   \\
LR Warmup                                       & 0.01                                   \\
Gradient Clipping                               & 1.0                                    \\
Training Precision                              & BF16                                  \\
Activation Function                             & Swiglu
                            
\end{tabular}
\caption{Model and training hyperparameters for the 1.4B transformer models trained in our ablation experiments.}
\label{ablations_hyperparam_table}
\end{table}

For the Pythia comparisons, we trained on 300B tokens (the same size as full Pile), using the model and training hyperparameters reported in \citet{biderman2023pythia}.

All models were trained on Nvidia H100 DGX systems. Most ablations were run on 1-2 DGX nodes. We used the MegatronLM \citep{shoeybi2019megatron} framework for training the models used in these ablation studies.

\section{Societal impacts}
We release Zyda with the aim of providing a dataset to make language model training at the trillion-token scale more accessible and effective. Empowering disparate and less-resourced groups to create such language models may have a variety of societal impacts. While enabling more actors to train effective language models may lead to greater decentralization of power and more equal access to the technology, it also opens the door to potential misuse. While we have endeavoured with our filtering to remove harmful and personally identifying content from Zyda, we cannot guarantee that Zyda cannot be used to train language models which could cause harm.

\newpage

\section{Ablation performance by evals}

\begin{figure*}[!htb] % the "" makes the figure span both columns
    \centering
    \begin{subfigure}{0.45\textwidth}
        \includegraphics[width=\textwidth]{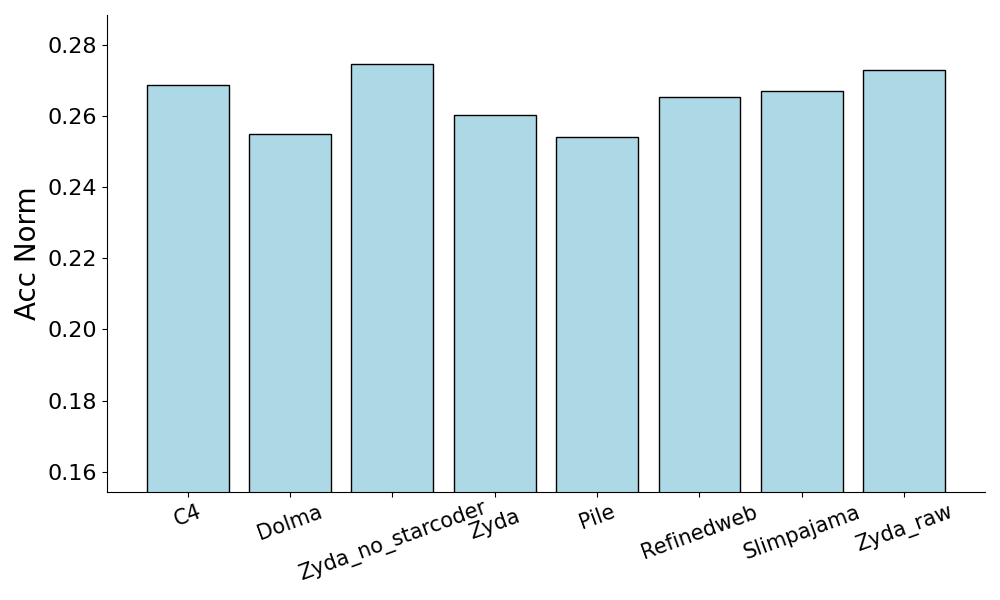}
        \caption{ARC Challenge}
    \end{subfigure}
    \hfill
    \begin{subfigure}{0.45\textwidth}
        \includegraphics[width=\textwidth]{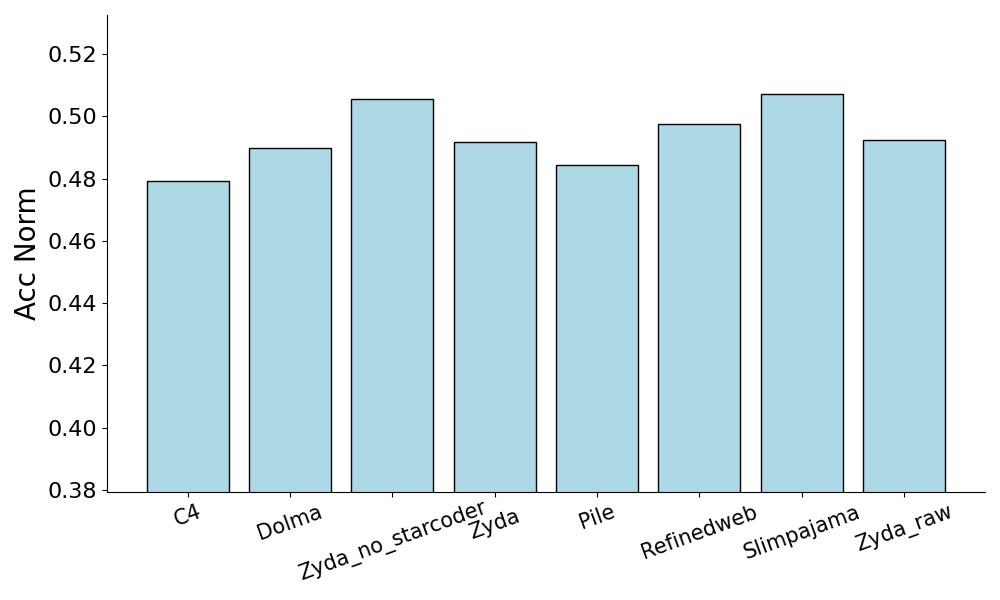}
        \caption{ARC Easy}
    \end{subfigure}
    \\
    \begin{subfigure}{0.45\textwidth}
        \includegraphics[width=\textwidth]{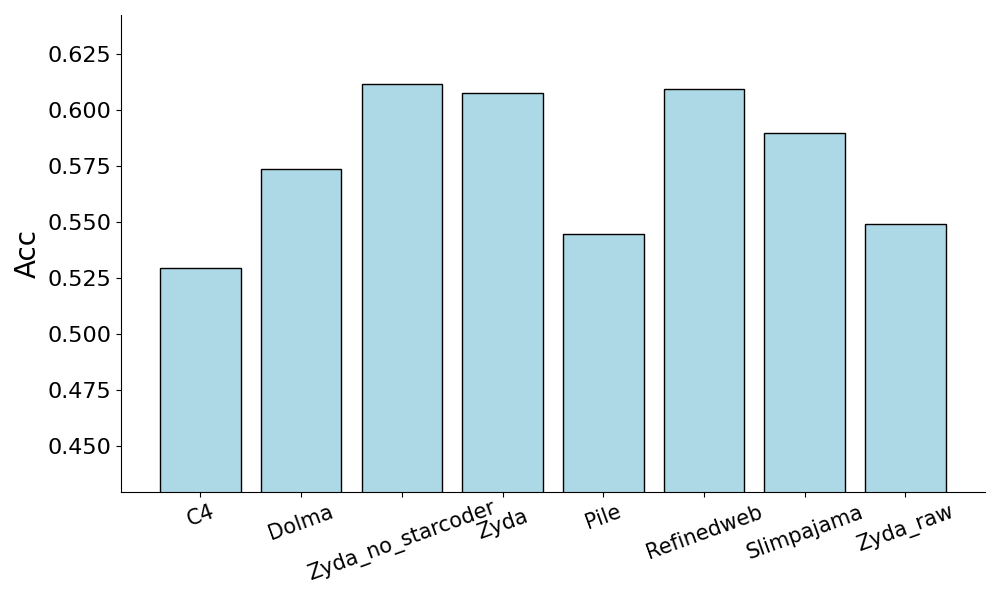}
        \caption{BoolQ}
    \end{subfigure}
    \hfill
    \begin{subfigure}{0.45\textwidth}
        \includegraphics[width=\textwidth]{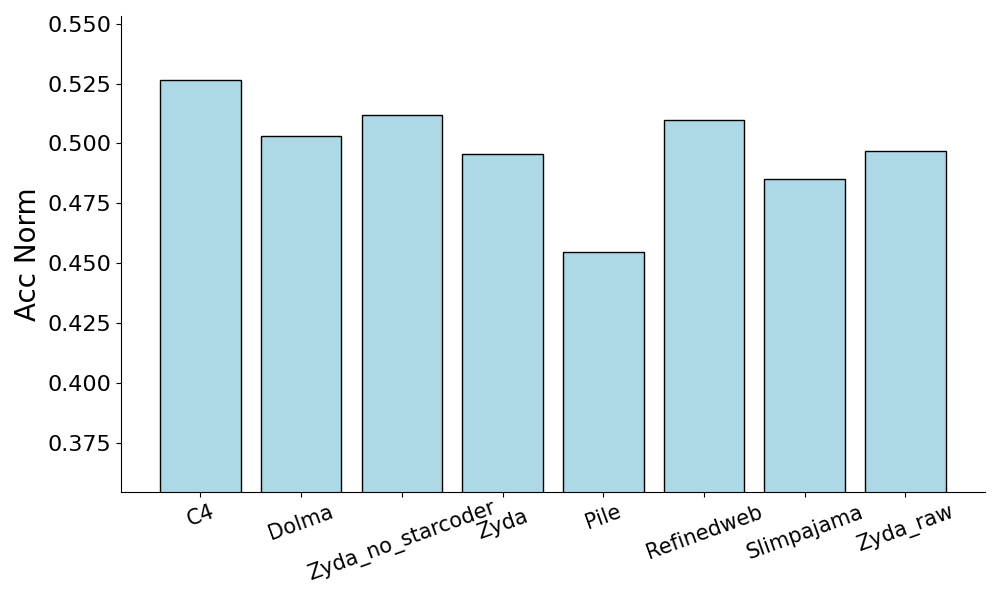}
        \caption{HellaSwag}
    \end{subfigure}
    \\
    \begin{subfigure}{0.45\textwidth}
        \includegraphics[width=\textwidth]{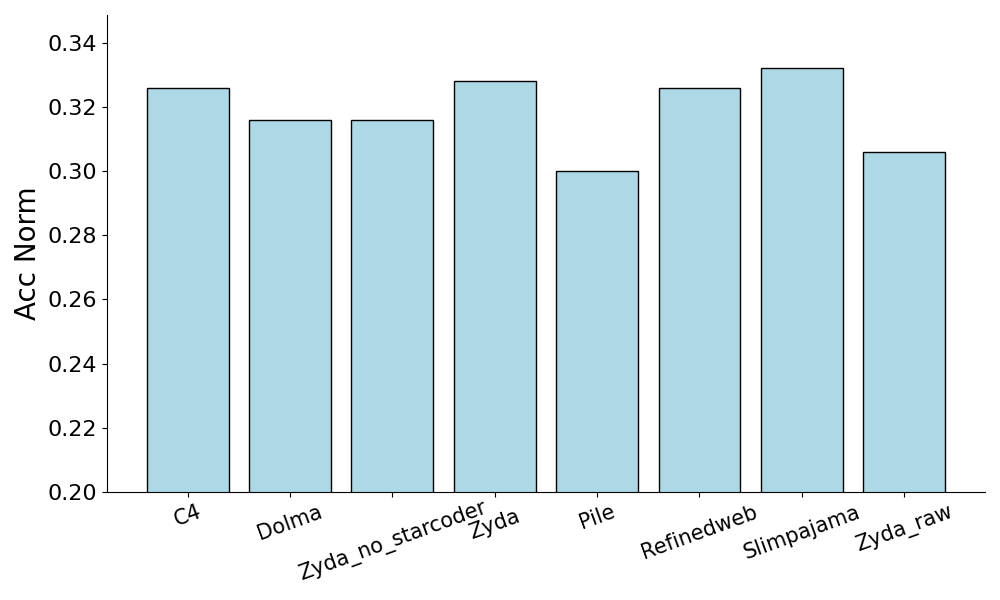}
        \caption{OpenBookQA}
    \end{subfigure}
    \hfill
    \begin{subfigure}{0.45\textwidth}
        \includegraphics[width=\textwidth]{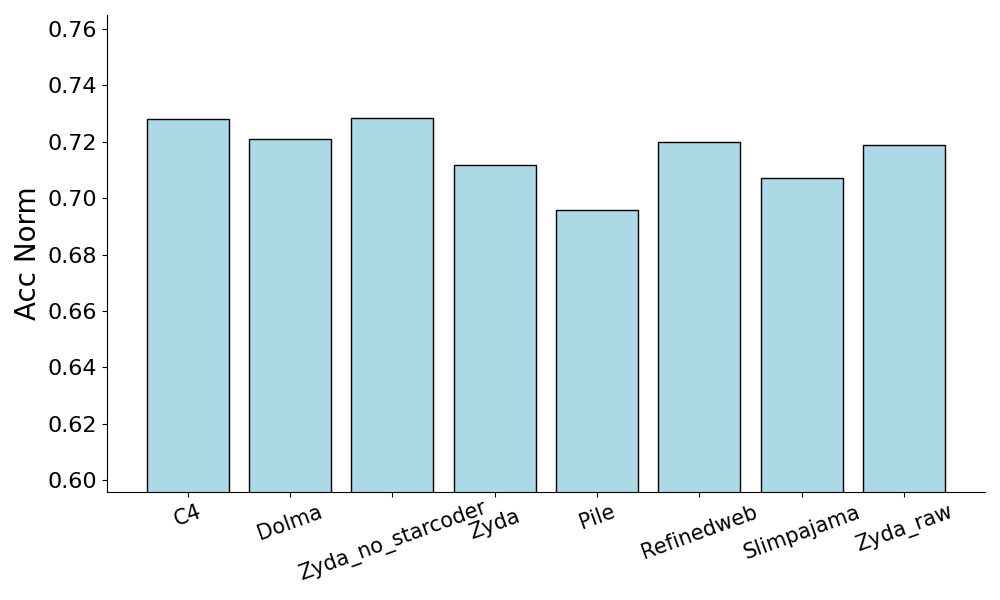}
        \caption{PIQA}
    \end{subfigure}
    \\
    \begin{subfigure}{0.45\textwidth}
        \includegraphics[width=\textwidth]{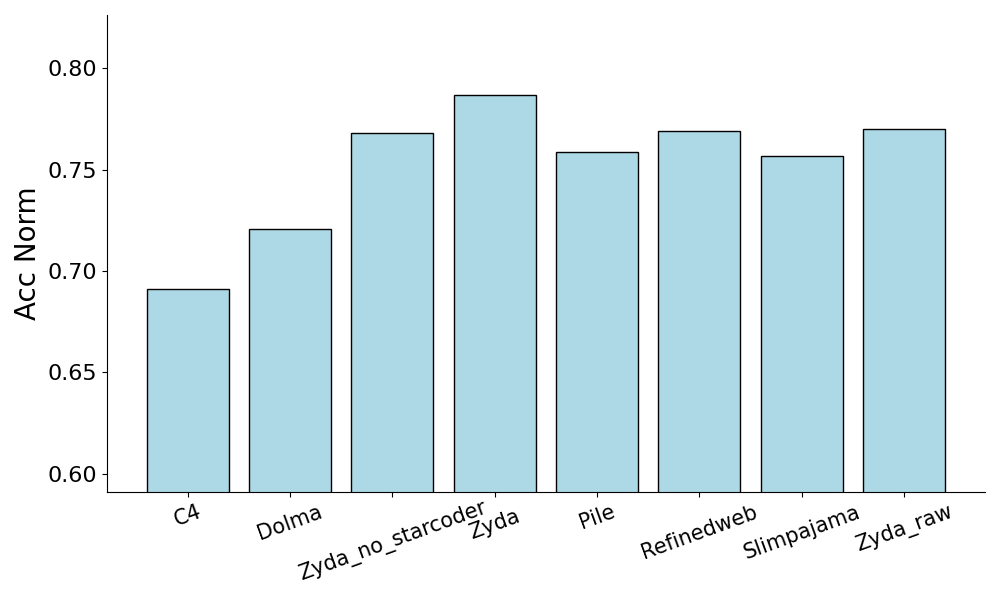}
        \caption{SciQ}
    \end{subfigure}
    \hfill
    \begin{subfigure}{0.45\textwidth}
        \includegraphics[width=\textwidth]{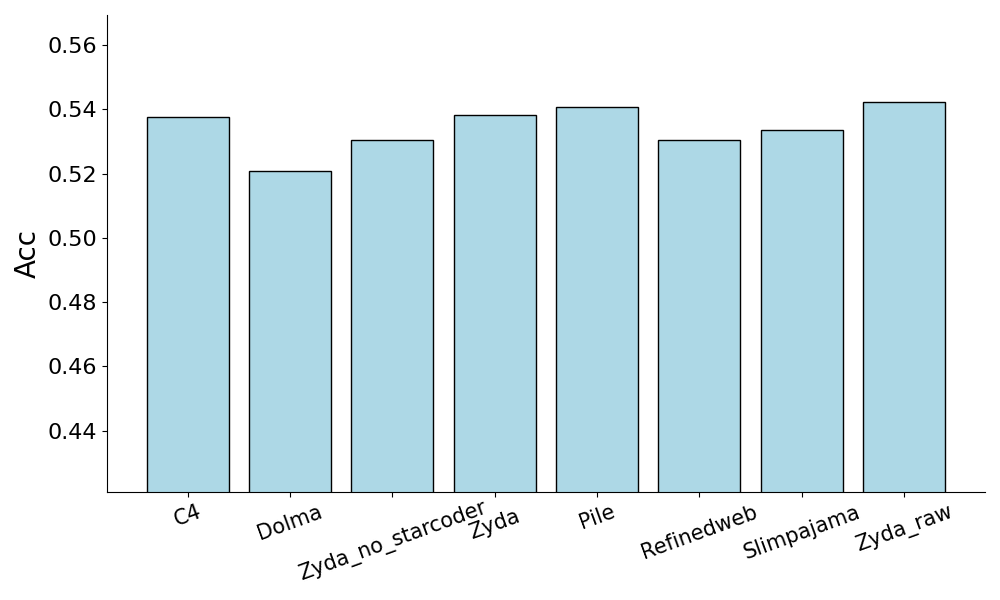}
        \caption{WinoGrande}
    \end{subfigure}
    \caption{Scores broken out by specific evaluation of our ablation studies comparing Zyda to alternative datasets. }
\end{figure*}

\newpage

\section{Additional details for dataset processing}
\subsection{Additional Filtering Details}
\label{Filtering_details}
In this Appendix, we describe the filters that were implemented in our filtering pipeline as well as additional filters that we experimented with which we did not find useful. We also present a number of qualitative insights we found while exploring Zyda in depth during our manual analysis and tuning of the filters.

The filters that were used in the final pipeline were as follows:
\begin{itemize}
    \item \textbf{Filter by min length}: This filtered out all documents shorter than a given length (we choose 100 characters). In practice, almost all documents shorter than this length consisted of short strings of broken formatting, ads and spam, single links, or unintelligble broken code snippets (primarily client-side javascript and sometimes CSS). This filter appeared highly effective with few false positives.
    \item \textbf{Filtering by max mean word length}: This filter removes documents which have too large a proportion of words of exceptional length. We found that this primarily removed short code snippets with verbose variables in CamelCase as well as malformed documents without spaces or with exceptionally long unusual strings such as hashes, cryptographic keys, and general weirdness.
    \item \textbf{Filtering by minimum mean word length}: This filter removed documents which too large a proportion of words with too short a length. We found in practice that this tended to remove short documents which were 'broken' in some sense -- usually consisting of either individual random characters or short bursts of random gibberish characters.
    \item \textbf{Filtering by fraction of alphanumeric characters}: This filter primarily removed documents consisting of non-alphanumeric characters. In practice, almost all such documents appeared to be gibberish with small fractions of unobjectionable code as false positives. We do not believe that this filter is suitable for full code datasets.
    \item \textbf{Filter by fraction numerical}: This filter removed documents consisting primarily of numbers. We detected a large number of such documents, most of which appeared to be stored data from various souces, most without any kind of context which would make it easy for a language model to understand what the data is meant to represent.
    \item \textbf{Filtering by fraction `xml'}: This filter removed egregious xml documents which primarily appeared to be endless sequences of xml tags with little actual content.
    \item \textbf{Filter by lorem ipsum}: This filter removed lorem ipsum text, of which we found a small amount in the Pile.
    \item \textbf{Filter by `http://' and `https://' and `www.'}: this filter removed pages with too high a fraction of website links. In practice, this removed documents which either consisted entirely of a single random URL (which could be long enough to survive our min length filter), or else documents with large proportions of URLs. Such documents were almost entirely ads or spam of various kinds, or else isolated links bars from websites. This filter appeared highly effective at removing such content in practice.
    \item \textbf{Filter by fraction `<' and `>'}: This filter removed a small amounts of documents consisting almost entirely of these symbols, as well as many poorly parsed webpages with truly excessive amounts of HTML tags without much appreciable content.
    \item \textbf{Filtering by `\":'}: This was used to filter for the opening of a python dictionary or JSON key or value. We used this to remove documents that were composed of an excessive fraction of JSON data. We found that in practice such documents primarily consisted of either unintelligble data or extremely repetitive CSS UI attributes.
    \item \textbf{Filtering by pornographic word list}: We found that using a relatively simple word list filtered out a large quantity of pornographic content from the Pile. This mostly consisted of adverts or titles from porn sites with a dash of porn ads. It also filtered out a fair amount of erotica fiction which may not necessarily be low quality. After experimenting with existing approaches, we elected to create a custom porn filter based only on a few keywords which can run fast through the dataset. We then filter both on the absolute number of such terms appearing in the document as well as on the fraction of the document comprised of these terms. We found that approximately $0.5\%$ of The Pile appears to be pornographic content.
    \item \textbf{Filter by profanity word list}: This filtered out documents based on a high proportion of common (English) profanity. We found that this primarily filtered out short documents consisting of low quality reddit comments or other social media postings. Notably, almost all documents removed by this filter were extremely short.
    \item \textbf{Filter by Chinese porn signals}: This filter was derived from \citet{kudugunta2024madlad} and used a special word-list that they had derived. Upon inspection this did appear to filter out mostly chinese text and included very little English. However we did not analyze this in depth as we focused on English monolingual language modelling. 
    \item \textbf{Filter by cursed string fraction}: We filtered using the ‘cursed regex’ of \citet{kudugunta2024madlad} which they claimed captured many undesirable strings. We found that this primarily filtered for ads, spam, and lorem ipsum text. This filter required quite a bit of manual tuning as the settings described in the paper were too aggressive – e.g. removing any line with the word ‘Facebook’, or ‘download’ – which clearly matches to many perfectly unobjectionable texts. 
 
\end{itemize}

There was also a number of filters that we tried and found ineffective, including a number of recommendations from prior works. These include:

\begin{itemize}
    \item \textbf{Filtering by max length}: We found that this was actively unhelpful and that almost all long documents appeared high quality. Usually they were either full books, or else concatenated encyclopedia entries or legal cases. We suspect that such data is actually high quality and helpful for the model.
    \item \textbf{Filtering by fraction of upper case characters}, as suggested in RedPajama. We found that while this filter caught some spam it also caught large amounts of unobjectionable text. Such text included notices and long-form advertisements from news papers, legal texts, and text which just seemed to be randomly have been converted to upper case. Since we could not find a threshold which had a sufficiently low false-positive level while also filtering out an appreciable number of documents, we ended up dropping this filter.
    \item We found that removing all documents with many “\{“ as done in \citet{kudugunta2024madlad} and with sentences ended without punctuation \cite{raffel2020exploring} primarily removed code which was otherwise non-objectionable.
    \item We found that removing documents without at least 2 standard english words such as “the, of” etc as proposed in \citet{rae2021scaling} tended to remove large fractions of otherwise unobjectionable code
    \item We found that filtering based on a large number of repeated characters was not effective and had many false positives. This usually meant either texts with many 0s in it – for instance numbers reported to a high degree of precision, and good text just with messed up formatting – i.e. many repeating dashes or slashes or tabs. There were not many files with extremely large amounts of repeating characters that we found. For instance, we found many legal cases with extremely large numbers of dashes, tabs, carriage-returns and \\xa0 characters which we filtered based on. Instead we opted to replace many repeated characters with a much smaller number of the same character. We hope this fixed many elements of the formatting without removing whole documents of good text which just happened to have been badly formatted during text extraction.
    \item We found that filtering by PII count was not particularly useful. We found that filtering by IP addresses as done in RefinedWeb \citep{penedo2023refinedweb} primarily removes tutorial networking code using example IP addresses which seems likely to be helpful for the model. Filtering aggressively by email and phone numbers tends to remove large numbers of otherwise useful social media and blog postings, however less aggressive filters which require a large fraction of such PII does remove some ads and spam.
    %\item We found that filtering based on a large number of repeated characters was not effective and had many false positives. This usually meant either texts with many 0s in it – for instance mathematical and scientific texts with numbers reported to a high degree of precision, as well as good text with broken formatting such as many repeating dashes or slashes or tabs. There were not many files with extremely large amounts of repeating characters that we found which seemed to be of intrinsically poor quality. For instance, we found many legal cases with extremely large numbers of dashes, tabs and carriage-return characters which we filtered based on. Instead we opted to replace many repeated characters with a much smaller number of the same character. We hope this fixes many elements of the formatting without trashing whole documents of good text just with bad formatting.
\end{itemize}

\subsection{Number of documents removed by each filter per dataset}

\begin{table*}[htbp]
\centering
\begin{tabular}{|>{\columncolor[HTML]{EFEFEF}}l|r|}
\hline
\cellcolor[HTML]{C0C0C0}\textbf{\makebox[8cm]{Filter}} & \cellcolor[HTML]{C0C0C0}\textbf{\makebox[4cm]{Num Documents Removed}} \\ \hline
\code{min\_mean\_word\_length} & 16,114,952 \\ \hline
\code{min\_length} & 1,130,814 \\ \hline
\code{max\_fraction} for pattern \textbf{https://} & 1,026,516 \\ \hline
\code{max\_fraction\_numerical} & 826,063 \\ \hline
\code{max\_fraction} for pattern \textbf{<} & 579,718 \\ \hline
\code{max\_fraction} for pattern \textbf{":} & 566,340 \\ \hline
\code{max\_PII\_items\_count} & 308,965 \\ \hline
\code{max\_fraction\_non\_alphanumeric} & 300,589 \\ \hline
\code{max\_fraction} for word list \textbf{sexual\_word\_list.json} & 219,983 \\ \hline
\code{max\_mean\_word\_length} & 159,312 \\ \hline
\code{max\_count} for pattern \textbf{xml} & 132,400 \\ \hline
\code{max\_count} for word list \textbf{sexual\_word\_list.json} & 42,030 \\ \hline
\code{max\_count} for pattern \textbf{<?xml version=} & 41,449 \\ \hline
\code{max\_fraction} for word list \textbf{cursed\_substrings.json} & 29,740 \\ \hline
\code{max\_fraction} for word list \textbf{profanity\_word\_list.json} & 26,419 \\ \hline
\code{max\_repeated\_substrings} & 11,705 \\ \hline
\code{max\_fraction} for pattern \textbf{www.} & 9,173 \\ \hline
\code{max\_count} for word list \textbf{zh\_pornsignals.json} & 8,419 \\ \hline
\code{max\_count} for pattern \textbf{lorem ipsum} & 1,534 \\ \hline
\end{tabular}
\caption{Filtering counts for pile-uncopyrighted}
\label{tab:filter-pile}
\end{table*}
\newpage
\vspace{5cm}

\begin{table*}[htbp]
\centering
\begin{tabular}{|>{\columncolor[HTML]{EFEFEF}}l|r|}
\hline
\cellcolor[HTML]{C0C0C0}\textbf{\makebox[8cm]{Filter}} & \cellcolor[HTML]{C0C0C0}\textbf{\makebox[4cm]{Num Documents Removed}} \\ \hline
\code{min\_length} & 2,090,400 \\
\code{min\_mean\_word\_length} & 570,427 \\
\code{max\_fraction} for pattern \textbf{https://} & 537,657 \\
\code{max\_fraction} for word list \textbf{cursed\_substrings.json} & 361,818 \\
\code{max\_mean\_word\_length} & 288,769 \\
\code{max\_fraction\_numerical} & 206,738 \\
\code{max\_PII\_items\_count} & 79,543 \\
\code{max\_fraction} for word list \textbf{profanity\_word\_list.json} & 34,400 \\
\code{max\_fraction} for pattern \textbf{":} & 23,955 \\
\code{max\_fraction} for pattern \textbf{www.} & 15,789 \\
\code{max\_fraction\_non\_alphanumeric} & 13,111 \\
\code{max\_count} for pattern \textbf{xml} & 3,411 \\
\code{max\_count} for word list \textbf{zh\_pornsignals.json} & 2,933 \\
\code{max\_fraction} for word list \textbf{sexual\_word\_list.json} & 2,528 \\
\code{max\_fraction} for pattern \textbf{<} & 2,512 \\
\code{max\_repeated\_substrings} & 418 \\
\code{max\_count} for pattern \textbf{<?xml version=} & 86 \\
\code{max\_count} for word list \textbf{sexual\_word\_list.json} & 2 \\
\hline
\end{tabular}
\vspace{1ex}
\caption{Filtering counts for C4-EN}
\label{tab:filter-c4}
\end{table*}
\vspace{5cm}
\newpage

\begin{table*}[htbp]
\centering
\begin{tabular}{|>{\columncolor[HTML]{EFEFEF}}l|r|}
\hline
\cellcolor[HTML]{C0C0C0}\textbf{\makebox[8cm]{Filter}} & \cellcolor[HTML]{C0C0C0}\textbf{\makebox[4cm]{Num Documents Removed}} \\ \hline
\code{max\_fraction} for word list \textbf{sexual\_word\_list.json} & 19,749 \\
\code{max\_PII\_items\_count} & 7,570 \\
\code{min\_mean\_word\_length} & 7,366 \\
\code{max\_count} for word list \textbf{sexual\_word\_list.json} & 4,833 \\
\code{max\_fraction} for pattern \textbf{https://} & 1,314 \\
\code{max\_fraction} for pattern \textbf{<} & 1,076 \\
\code{max\_fraction} for pattern \textbf{":} & 698 \\
\code{max\_fraction} for word list \textbf{cursed\_substrings.json} & 669 \\
\code{max\_repeated\_substrings} & 205 \\
\code{max\_count} for pattern \textbf{xml} & 151 \\
\code{max\_mean\_word\_length} & 139 \\
\code{max\_count} for word list \textbf{zh\_pornsignals.json} & 137 \\
\code{max\_count} for pattern \textbf{<?xml version=} & 25 \\
\code{max\_fraction} for word list \textbf{profanity\_word\_list.json} & 21 \\
\code{max\_fraction\_numerical} & 17 \\
\code{max\_fraction\_non\_alphanumeric} & 6 \\
\code{max\_count} for pattern \textbf{lorem ipsum} & 2 \\
\hline
\end{tabular}
\vspace{1ex}
\caption{Filtering counts for peS2o}
\label{tab:filter-pes2o}
\end{table*}
\vspace{5cm}
\newpage

\begin{table*}[htbp]
\centering
\begin{tabular}{|>{\columncolor[HTML]{EFEFEF}}l|r|}
\hline
\cellcolor[HTML]{C0C0C0}\textbf{\makebox[8cm]{Filter}} & \cellcolor[HTML]{C0C0C0}\textbf{\makebox[4cm]{Num Documents Removed}} \\ \hline
\code{min\_mean\_word\_length} & 175,161 \\
\code{max\_PII\_items\_count} & 11,901 \\
\code{max\_repeated\_substrings} & 752 \\
\code{max\_fraction} for word list \textbf{cursed\_substrings.json} & 615 \\
\code{max\_fraction\_numerical} & 502 \\
\code{max\_count} for pattern \textbf{xml} & 166 \\
\code{max\_count} for word list \textbf{zh\_pornsignals.json} & 113 \\
\code{max\_count} for word list \textbf{sexual\_word\_list.json} & 67 \\
\code{max\_fraction} for pattern \textbf{":} & 32 \\
\code{max\_count} for pattern \textbf{<?xml version=} & 29 \\
\code{max\_fraction} for pattern \textbf{https://} & 24 \\
\code{max\_mean\_word\_length} & 18 \\
\code{max\_fraction\_non\_alphanumeric} & 15 \\
\code{min\_length} & 2 \\
\code{max\_count} for pattern \textbf{lorem ipsum} & 2 \\
\code{max\_fraction} for word list \textbf{sexual\_word\_list.json} & 1 \\
\hline
\end{tabular}
\caption{Filtering counts for arXiv}
\label{tab:filter-arxiv}
\end{table*}
\newpage
\vspace{5cm}

\begin{table*}[htbp]
\centering
\begin{tabular}{|>{\columncolor[HTML]{EFEFEF}}l|r|}
\hline
\cellcolor[HTML]{C0C0C0}\textbf{\makebox[8cm]{Filter}} & \cellcolor[HTML]{C0C0C0}\textbf{\makebox[4cm]{Num Documents Removed}} \\ \hline
\code{min\_length} & 11,527,798 \\
\code{min\_mean\_word\_length} & 3,085,862 \\
\code{max\_fraction} for word list \textbf{sexual\_word\_list.json} & 2,037,579 \\
\code{max\_fraction\_numerical} & 1,436,988 \\
\code{max\_PII\_items\_count} & 826,732 \\
\code{max\_fraction} for word list \textbf{cursed\_substrings.json} & 794,444 \\
\code{max\_count} for word list \textbf{sexual\_word\_list.json} & 492,413 \\
\code{max\_count} for word list \textbf{zh\_pornsignals.json} & 357,466 \\
\code{max\_fraction} for pattern \textbf{":} & 286,229 \\
\code{max\_fraction} for word list \textbf{profanity\_word\_list.json} & 260,606 \\
\code{max\_fraction for pattern} < & 140,598 \\
\code{max\_mean\_word\_length} & 28,146 \\
\code{max\_fraction\_non\_alphanumeric} & 28,036 \\
\code{max\_count} for pattern \textbf{xml} & 24,755 \\
\code{max\_fraction} for pattern \textbf{https://} & 7,807 \\
\code{max\_count} for pattern \textbf{<?xml version=} & 7,381 \\
\code{max\_repeated\_substrings} & 3,090 \\
\code{max\_fraction} for pattern \textbf{www.} & 2,686 \\
\code{max\_count} for pattern \textbf{lorem ipsum} & 1,061 \\
\hline
\end{tabular}
\caption{Filtering counts for RefinedWeb}
\label{tab:filter-refinedweb}
\end{table*}
\newpage
\vspace{5cm}

\begin{table*}[htbp]
\centering
\begin{tabular}{|>{\columncolor[HTML]{EFEFEF}}l|r|}
\hline
\cellcolor[HTML]{C0C0C0}\textbf{\makebox[8cm]{Filter}} & \cellcolor[HTML]{C0C0C0}\textbf{\makebox[4cm]{Num Documents Removed}} \\ \hline
\code{min\_mean\_word\_length} & 14,682,212 \\ \hline
\code{max\_fraction} for pattern \textbf{https://} & 1,808,881 \\ \hline
\code{min\_length} & 847,055 \\ \hline
\code{max\_fraction} for pattern \textbf{<} & 704,490 \\ \hline
\code{max\_fraction\_numerical} & 689,546 \\ \hline
\code{max\_fraction} for pattern \textbf{":} & 66,8174 \\ \hline
\code{max\_PII\_items\_count} & 570,952 \\ \hline
\code{max\_mean\_word\_length} & 348,773 \\ \hline
\code{max\_fraction} for word list \code{cursed\_substrings.json} & 286,527 \\ \hline
\code{max\_count} for pattern \textbf{xml} & 143,725 \\ \hline
\code{max\_fraction\_non\_alphanumeric} & 112,351 \\ \hline
\code{max\_count} for pattern \textbf{<?xml version=} & 43,069 \\ \hline
\code{max\_count} for word list \textbf{sexual\_word\_list.json} & 40,855 \\ \hline
\code{max\_fraction} for word list \textbf{sexual\_word\_list.json} & 30,420 \\ \hline
\code{max\_fraction} for word list \textbf{profanity\_word\_list.json} & 28,863 \\ \hline
\code{max\_fraction} for pattern \textbf{www.} & 18,936 \\ \hline
\code{max\_count} for word list \textbf{zh\_pornsignals.json} & 10,128 \\ \hline
\code{max\_repeated\_substrings} & 6,555 \\ \hline
\code{max\_count} for pattern \textbf{lorem ipsum} & 3,231 \\ \hline
\end{tabular}
\caption{Filtering counts for SlimPajama}
\label{tab:filter-slimpj}
\end{table*}

\newpage
% \FloatBarrier

\subsection{Duplicates sources}

\begin{figure*}[!htb]
    \centering
    \includegraphics[width=\textwidth]{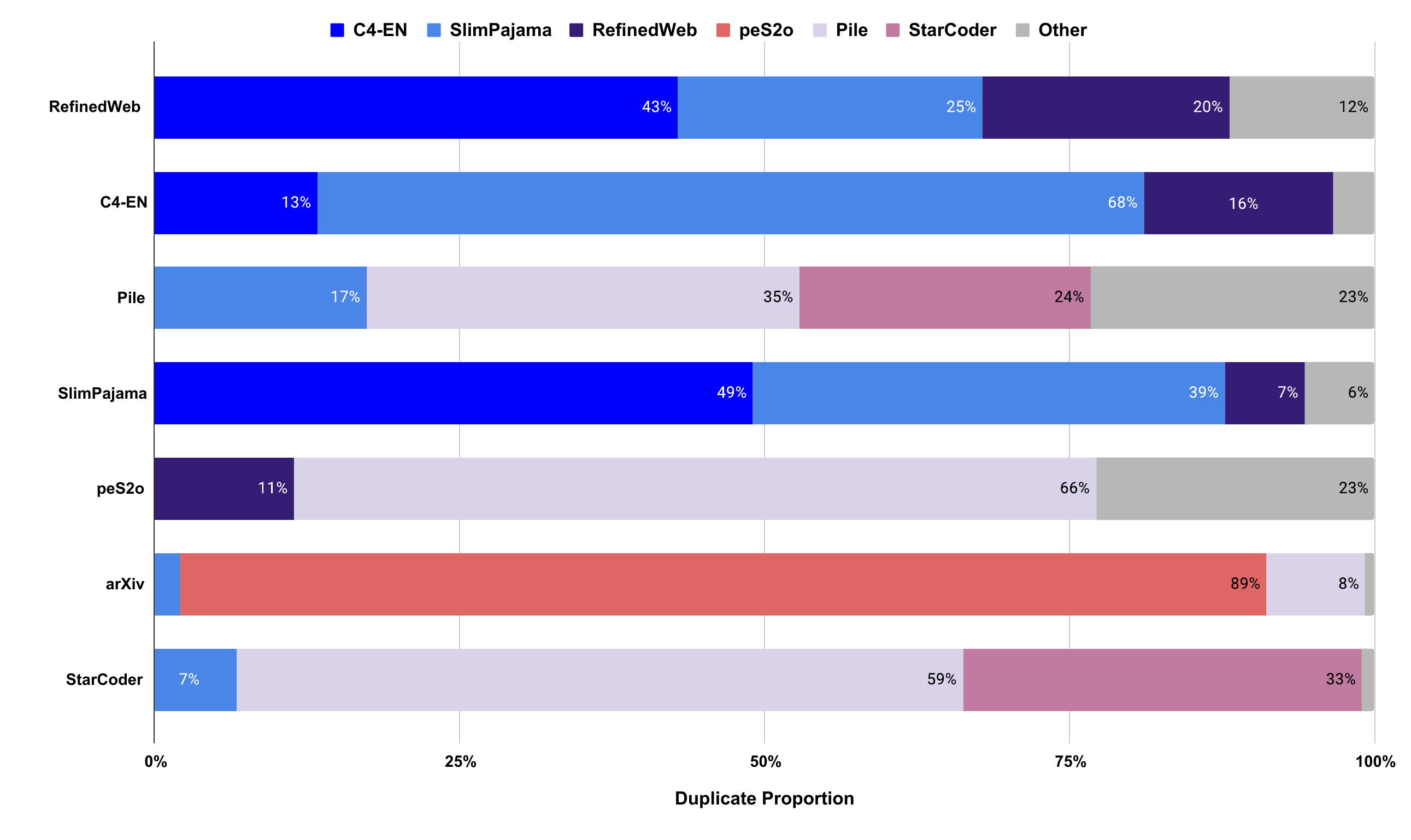}
    \vspace{-0cm}
    \caption{Proportion of duplications for all datasets for LSH 40\%.}
    \label{fig:duplication-proportions}
\end{figure*}

We computed the distribution of sources of duplicates across the components (see Figure \ref{fig:duplication-proportions}). On Figure 7 we plot the proportion of duplicates that are coming from different datasets: e.g., for Pile we identified roughly 90 million documents duplicated in the Pile itself and other datasets, and the figure tells us that 35\% of those are coming from itself, 24\% from StarCoder, 17\% from SlimPajama, and 23\% from the rest of the components. We observe extremely high number of duplicates in arXiv are coming from peS2o since they have the same provenance. We also observe that our approach correctly identified a lot of documents in SlimPajama are present in C4, since C4 was included in RedPajama (the parent dataset of SlimPajama). Big proportion of duplicates in RefinedWeb, C4 overlap with each other probably due the fact that they are derivative of Common Crawl.

%\end{appendices}
\newpage

\subsection{Estimating empirical false positive rate}

\begin{figure*}[!htb] % the "" makes the figure span both columns
    \centering
    \begin{subfigure}{0.45\textwidth}
        \raisebox{8ex}{\includegraphics[width=\textwidth]{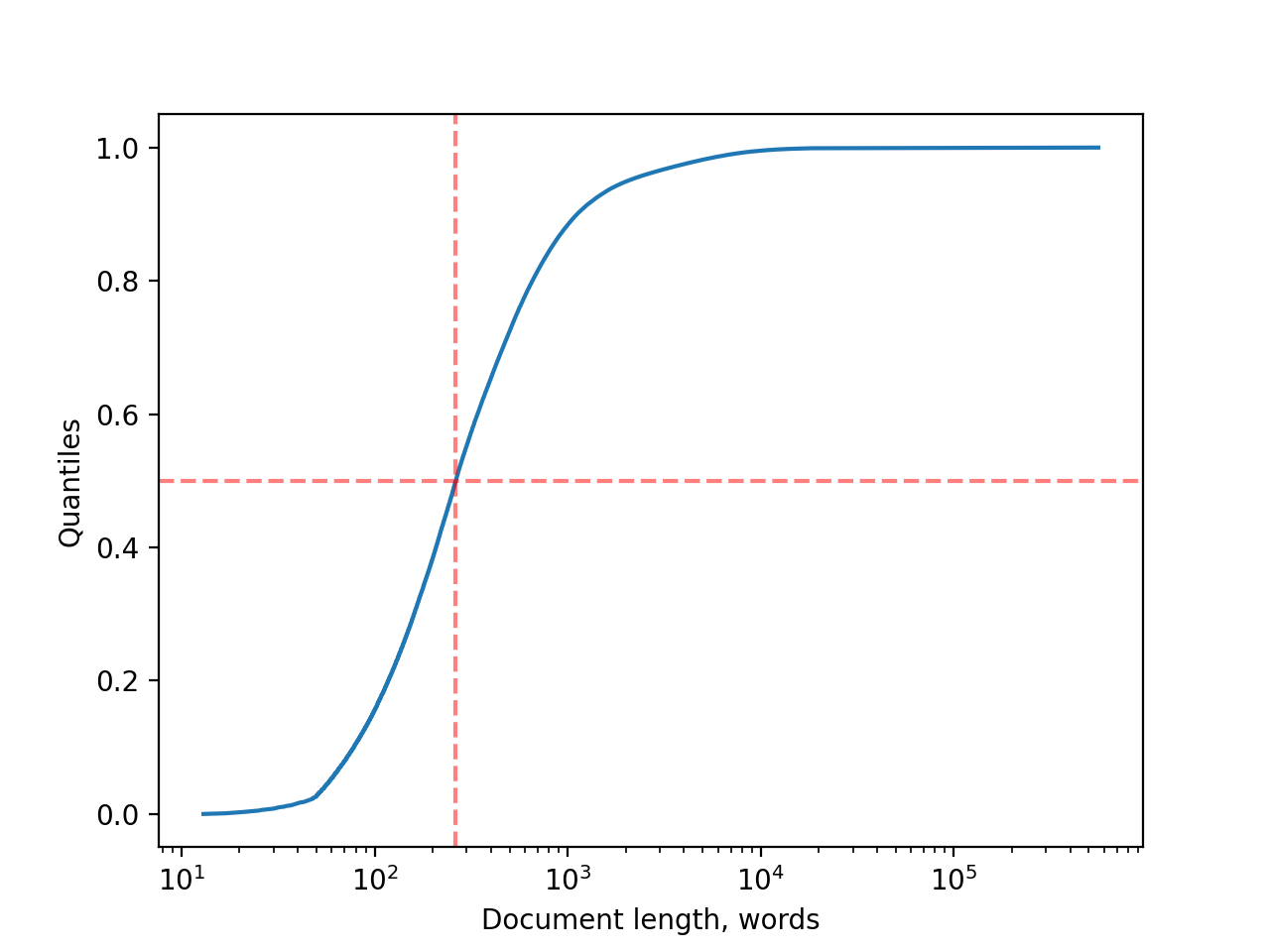}}
        \vspace{-6ex}
        \caption{Cumulative distribution function of duplicate documents lengths on a log scale. Red dashed lines correspond to median length. We observe that the vast majority of documents tend to sit between $10^2$ and $10^4$ words in length.}
        \label{fig:dup-len}
    \end{subfigure}
    \hfill
    \begin{subfigure}{0.45\textwidth}
        \includegraphics[width=\textwidth]{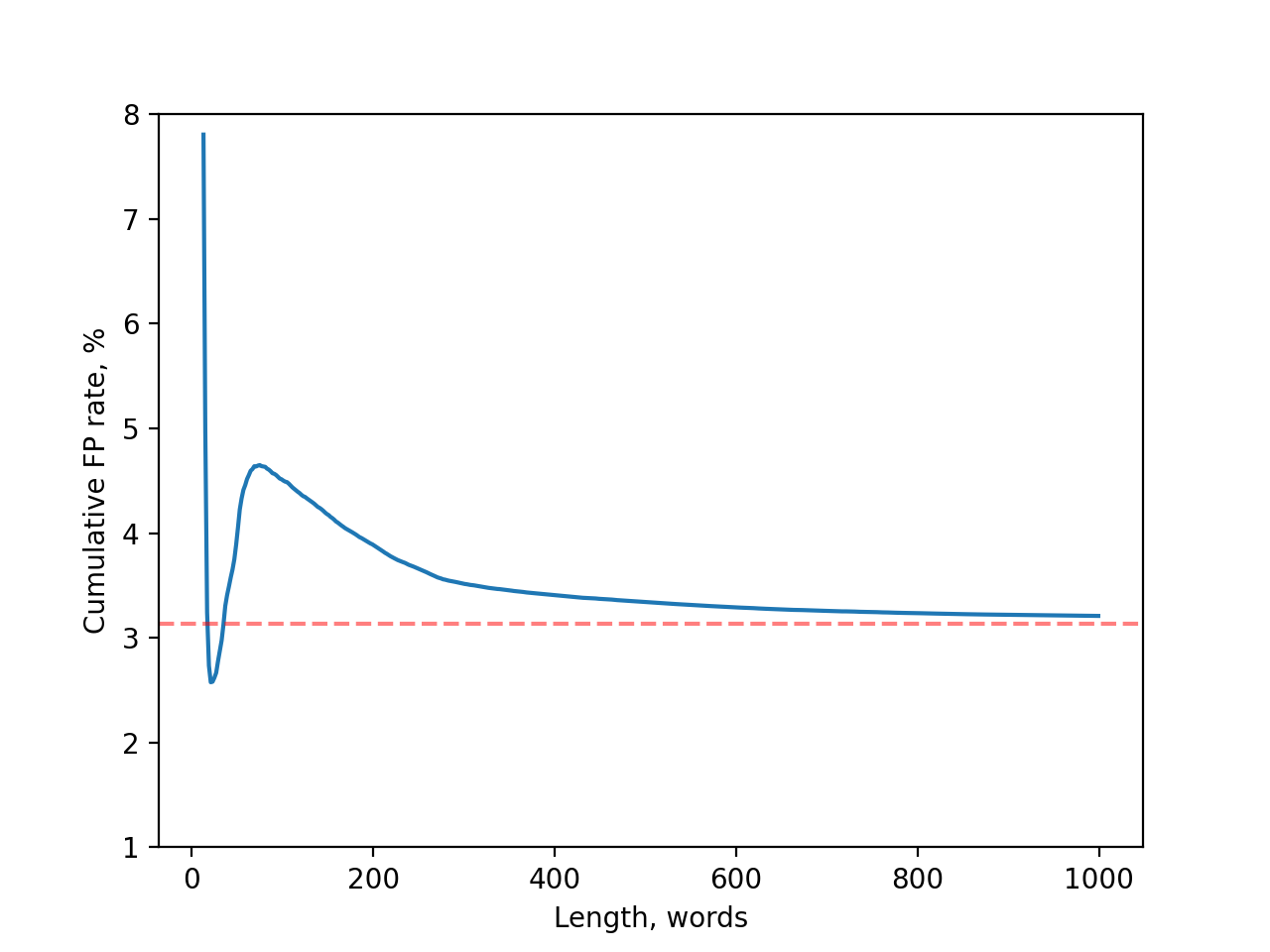}
        \caption{Cumulative FP rate as a function of duplicate document length. The red dashed line corresponds to the theoretical 3.1\% FP rate computed on all pairs of the sample. We observe that for short documents the empirical and theoretical FP rates diverge before converging to the theoretical prediction for long documents}
        \label{fig:FP-len}
    \end{subfigure}
    \caption{Statistics of the sample of 4.8 million duplicate documents.}
\end{figure*}

To estimate the true false positive rate beyond the theoretical calculation, we sampled 4.8 million duplicates pairs (see Figure \ref{fig:dup-len} for the dustribution of lengths), and computed their empirical edit and Jaccard similarities, as well as manually inspected a number of examples. On Figure \ref{fig:FP-len} we plot a dependence on document length of what we call a cumulative FP rate: given length $l$, cumulative FP rate is the FP rate amoung duplicates for which the length is at most $l$. We found that the length of the document had a strong effect on the proportion of false positives with short documents being more likely to be categorized as false positives than longer ones. This is likely due to our use of 13-grams for similarity matching as well as the fact that noise fluctuations in similarity are of large scale relative to the threshold at small lengths.

%%%%%%%%%%%%%%%%%%%%%%%%%%%%%%%%%%%%%%%%%%%%%%%%%%%%%%%%%%%%

\end{document}